\documentclass[letterpaper, 10 pt, conference]{ieeeconf}  

\IEEEoverridecommandlockouts                              

\usepackage{amsmath} 
\usepackage{amssymb}  
\usepackage{graphicx}
\usepackage[table]{xcolor}
\usepackage{algorithm}
\usepackage{algpseudocode}
\algrenewcommand\algorithmicrequire{\textbf{Input:}}
\algrenewcommand\algorithmicensure{\textbf{Output:}}
\usepackage{CJKutf8}
\usepackage{url}
\usepackage[compatibility=false]{caption}
\usepackage[hidelinks]{hyperref}
\newcommand{\vect}[1]{\mathbf{#1}}

\newcommand{\dd}{\mathrm{d}}

\newcommand{\Cost}{\mathrm{Cost}}

\title{\LARGE \bf
Improving Robotic Imitation Learning via Trajectory Standardization
}

\author{Licheng Yang$^{1,2}$, Lingfeng Qian$^{2}$, Fei Zheng$^{2}$, Yonghao He$^{2\dagger}$, Wei Sui$^{2}$, Shuangshuang Li$^{1}$, and Hu Su$^{1*}$
\thanks{*Corresponding author.}
\thanks{$\dagger$Project lead.}%
\thanks{$^{1}$State Key Laboratory of Multimodal Artificial Intelligence Systems (MAIS), Institute of Automation, Chinese Academy of Sciences}%
\thanks{$^{2}$D-Robotics}%
}

\begin{document}
\hbadness=10000

\maketitle
\thispagestyle{empty}
\pagestyle{empty}

\begin{abstract}

Imitation learning for robotic manipulation relies on large sets of human demonstration trajectories, which are often noisy and temporally irregular due to variable operator speed, intermittent pauses, and inconsistent action density. A common preprocessing strategy is time-uniform downsampling to shorten sequences, but it cannot effectively remove speed-induced non-uniformity or redundant pauses. This mismatch degrades data quality and hinders policy learning. To address this issue, we propose Information-Standardized Trajectory Resampling (ISR), an offline preprocessing method for effective imitation learning. ISR resamples each trajectory by enforcing approximately equal information distance between adjacent points. Specifically, we map trajectories onto an information-modulated Riemannian manifold and perform geodesic-equidistant parameterization. We construct an information-intensity field from velocity and acceleration norms: the velocity term removes small-motion redundancy, while the acceleration term preserves high-curvature and fine-manipulation phases. We evaluate ISR on three real-world manipulation tasks with mainstream imitation learning policies. Compared with the baseline time-uniform $3\times$ downsampling, ISR improves task success rates by about $25\%$, remains robust across datasets collected from different operators, and reduces both dataset size and training cost. The code and videos are publicly available at \textcolor{blue}{\url{https://d-robotics-ai-lab.github.io/isr.page}}.

\end{abstract}

\section{Introduction}

Imitation learning (IL) enables robots to acquire manipulation skills from human demonstrations
and is a critical pathway toward general-purpose autonomy.
Because these methods learn directly from demonstration data,
the quality and consistency of demonstration datasets fundamentally affect the performance of learning policies.
In practice, however, teleoperated trajectories inevitably exhibit temporal non-uniformity
and spatial redundancy: variable operator speed, unintentional pauses,
and inconsistent action density introduce non-Markovian artifacts
and inflate dataset size without contributing useful information
\cite{tan2025think,zhao2023learning}.
Although recent generative policies—such as diffusion-based
\cite{chi2023diffusionpolicy,liu2024rdt,li2024cogact} and flow-matching approaches
\cite{zhang2025flowpolicy,intelligence2025pi_}—have
substantially improved the modeling of multimodal action distributions,
they remain sensitive to these data-quality issues;
noisy, redundant demonstrations still degrade learning efficiency and task success.
Despite its practical importance, trajectory standardization as a data-centric
preprocessing step has received limited attention in imitation learning.
In this work, we aim to design a resampling method that enforces
trajectory-level information consistency across demonstrations,
where adjacent resampled points carry comparable kinematic--dynamic information,
producing compact yet high-fidelity datasets that improve both training efficiency
and policy performance.

These data-quality problems manifest in two ways.
Within individual episodes, unintentional pauses and slow movements
at typical recording frequencies (30--60\,Hz)
flood the dataset with near-duplicate action points \cite{tan2025think}.
Under identical observations, the resulting ``stay'' and ``move'' commands
create non-Markovian contradictions \cite{zhao2023learning},
and history-conditioned policies trained on such sequences
tend to replay recent actions rather than reason about the task state
\cite{wen2020fighting}.
Across episodes, differences in operator pacing, style,
and skill level \cite{gandhi2023eliciting}
skew the data distribution:
frequent low-value motions dominate training gradients
while task-critical actions are underrepresented
\cite{parekh2025balancedbehaviorcloningimbalanced,orsini2021matters,mandlekar2021matters}.
Such redundancy also increases memory and latency costs for transformer-based policies with long contexts
\cite{kawaharazuka2025vla-survey}.

\begin{figure}[t]
  \centering
  \includegraphics[width=\columnwidth]{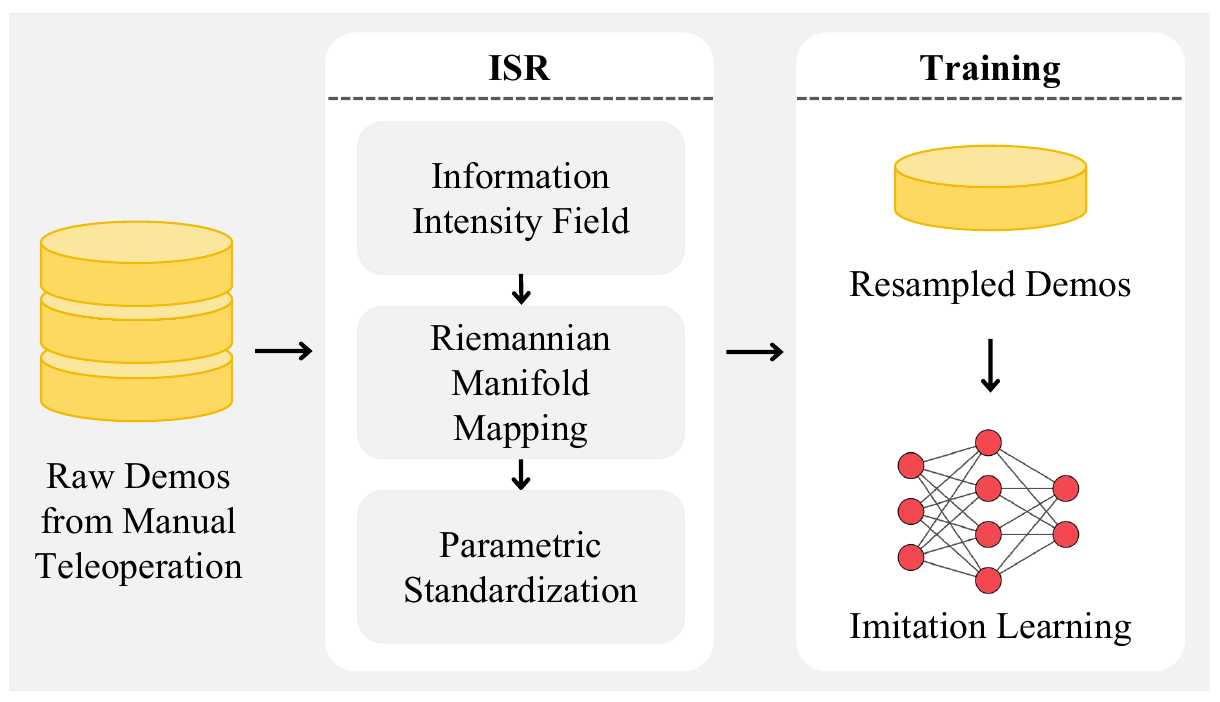}
  \caption{Overview of the ISR pipeline. Raw teleoperated demonstrations are resampled offline by ISR before being used to train an imitation learning policy.}
  \label{fig:overview}
\end{figure}
A common preprocessing step is to apply time-uniform downsampling
(typically $3\times$) to shorten high-frequency demonstration sequences
\cite{chi2023diffusionpolicy}.
While operationally simple, this strategy treats every time interval
as equally important and is blind to the kinematic and dynamic content
of the trajectory.
It cannot distinguish an unintentional pause from a deliberate deceleration
before fine manipulation, nor can it adapt sampling density to segments
where the end-effector undergoes rapid directional changes.
Alternatives based on Euclidean length
\cite{braglia2025arc,shi2023waypointbased} improve spatial uniformity
but remain insensitive to acceleration-level features
that encode the operator's dynamic intent.

We propose Information-Standardized Trajectory Resampling (ISR),
an offline preprocessing method that resamples demonstration trajectories
so that consecutive points carry approximately equal kinematic
and dynamic information.
The core idea is an information-intensity field constructed from
two complementary terms:
a \emph{velocity term} that measures the instantaneous displacement rate,
capturing first-order kinematic content;
and an \emph{acceleration term} that measures the rate of velocity change,
capturing second-order cues that serve as kinematic proxies
for force-sensitive intent, such as centripetal acceleration
through curved paths and tangential deceleration
before contact-rich manipulation phases.
Using this field as a conformal factor,
we map each trajectory onto a Riemannian manifold
and formulate resampling as a geodesic-equidistant optimization problem,
where consecutive resampled points are constrained
to have approximately equal generalized distance.
The velocity term compresses redundant low-displacement segments
such as pauses and slow transits,
while the acceleration term preserves dense sampling
through high-curvature transitions and force-critical phases.
The result is a compact, standardized trajectory that retains
the operator's manipulation intent with significantly fewer action points.

In summary, our contributions are:
\begin{itemize}
\item We identify trajectory standardization as an underexplored
yet practically important preprocessing problem in imitation learning,
and demonstrate that replacing time-uniform downsampling
with information-standardized resampling yields substantial gains
in policy performance.
\item We propose ISR, which constructs
a kinematic--dynamic information-intensity field
from velocity and acceleration norms,
maps trajectories onto the Riemannian manifold,
and solves geodesic-equidistant optimization
to produce standardized demonstrations.
As shown in Fig.~\ref{fig:overview},
ISR is a plug-and-play offline module
compatible with standard IL training pipelines.
\item We evaluate ISR with $\pi_{0.5}$ \cite{intelligence2025pi_}
and VO-DP \cite{ni2025vodp} on three real-world manipulation tasks.
Compared with time-uniform $3\times$ downsampling,
ISR consistently improves task success rates
while reducing dataset size and training cost,
and remains robust across datasets collected from different operators.
\end{itemize}

\section{Related Work}

Our work addresses trajectory-level data standardization
for imitation learning.
We review two related lines of research:
waypoint- and keyframe-based methods
that introduce sparse representations at the \emph{policy} level,
and trajectory simplification and resampling techniques
that operate at the \emph{data} level.

\subsection{Waypoint-Based and Keyframe-Based Learning}

Dense action sequences in imitation learning
are often accompanied by compounding errors under distributional shift
during closed-loop inference
\cite{ross2011reduction,zhao2023learning}.
Waypoint-based and keyframe-based methods mitigate this
by reducing the effective prediction horizon.
AWE  (Automatic Waypoint Extraction) \cite{shi2023waypointbased} extracts waypoints
from demonstrations via the Douglas--Peucker algorithm and replaces next-state prediction with next-waypoint prediction,
reducing the number of decisions the policy must make.
HYDRA \cite{belkhale2023hydra} takes a complementary approach
by introducing a hybrid action space
that pairs sparse high-level waypoints for free-space transit
with dense low-level actions for contact-rich manipulation.
VIEW \cite{jonnavittula2025view} extends waypoint extraction
to human video demonstrations,
compressing lengthy visual trajectories into sparse spatial targets
to bridge the morphological gap between humans and robots.
Beyond imitation learning,
waypoints also serve as sparse subgoal representations
in reinforcement learning \cite{mehta2024waypoint}
and as anchors for keyframe-based world-model generation
\cite{li2025keyworld}.

These methods modify the policy architecture or prediction target
to handle trajectory sparsity;
they do not standardize the underlying training data.
In contrast, ISR operates entirely at the data level:
it resamples the full demonstration trajectory
to achieve kinematic--dynamic information uniformity,
producing standardized datasets that are compatible
with any downstream policy without architectural changes.

\subsection{Trajectory Simplification and Resampling}

In practice, the most widely adopted data-level preprocessing
for imitation learning is time-uniform downsampling---typically $3\times$
on 30--60\,Hz recordings \cite{chi2023diffusionpolicy}.
While simple, it treats every time interval as equally important
and cannot adapt to varying information density along a trajectory.

Trajectory compression has been studied more broadly
in transportation and spatiotemporal data mining.
The Douglas--Peucker algorithm \cite{douglas1973algorithms}
is a classical geometric simplification method.
Its spatiotemporal extensions TD-TR (Top-Down Time Ratio) and OPW-TR (Opening Window Time Ratio)
\cite{meratnia2004spatiotemporal}
replace purely geometric error with time-ratio-based metrics,
but remain ineffective at filtering redundant pauses
typical of robotic teleoperation.
MDL (minimum description length)-based methods \cite{lee2007trajectory}
reformulate reconstruction error as a soft loss
for sub-trajectory partitioning,
but are designed for trajectory clustering
rather than for producing training data
with standardized kinematic properties.
In imitation learning, AWE \cite{shi2023waypointbased}
uses Douglas--Peucker for waypoint extraction,
but only changes the policy's prediction target;
the training dataset itself is not resampled or standardized.

Closest to our setting,
arc-length-based warping \cite{braglia2025arc}
re-parameterizes trajectories by Euclidean arc length
and extracts equidistant keypoints,
removing the need for temporal alignment
and mitigating variable-speed artifacts.
However, arc-length parameterization accounts only
for spatial displacement---a first-order kinematic quantity---and
cannot capture acceleration-level features
that encode dynamic intent.
More broadly, resampling as a principled data-standardization problem
remains largely unexplored in imitation learning,
despite the critical impact of trajectory quality
on downstream policy performance.

ISR addresses this gap by constructing
an information-intensity field from velocity
and acceleration norms,
moving trajectory resampling beyond purely spatial criteria
to a kinematic--dynamic information-standardization framework.

\section{Information-Standardized Resampling}

This section presents Information-Standardized Trajectory Resampling (ISR).
We begin with the problem definition,
introduce the information-intensity manifold,
formulate the geodesic-equidistant optimization,
and finally analyze the effects of the velocity
and acceleration cost terms.

\subsection{Problem Definition}

In imitation learning, demonstration trajectories are recorded
at high frequency (typically 30--60\,Hz) via teleoperation.
Let $\mathcal{T}_{\mathrm{raw}} = \{(\vect{p}_i,\, t_i)\}_{i=0}^{N-1}$
denote a raw trajectory, where $\vect{p}_i \in \mathbb{R}^3$
is the end-effector position at timestamp $t_i$.
From the position sequence we compute velocities and accelerations
via finite differences:
\begin{equation}
  \vect{v}_i = \frac{\vect{p}_{i+1} - \vect{p}_i}{\Delta t_i},
  \qquad
  \vect{a}_i = \frac{\vect{v}_i - \vect{v}_{i-1}}{\Delta t_i^{\mathrm{avg}}},
\end{equation}
where $\Delta t_i = t_{i+1} - t_i$ and
$\Delta t_i^{\mathrm{avg}} = (\Delta t_i + \Delta t_{i-1})/2$.
The velocity $\vect{v}_i$ describes the first-order kinematic state
of the end-effector---its instantaneous displacement rate and direction.
The acceleration $\vect{a}_i$ captures the second-order state:
although it does not measure contact force directly,
it provides a kinematic proxy for force-sensitive intent,
as operators express contact regulation through small decelerations,
alignments, and directional corrections that induce tangential
or centripetal acceleration despite limited displacement.
Taken together, velocity and acceleration provide a complete
kinematic--dynamic characterization of the information content
at each trajectory point.

Our objective is to extract a resampled subsequence
$\mathcal{K} = \{k_0, k_1, \ldots, k_{M-1}\} \subset \{0, 1, \ldots, N{-}1\}$,
with $0 = k_0 < k_1 < \cdots < k_{M-1} = N{-}1$
to preserve the start and end points,
such that adjacent resampled points are separated by approximately equal
\emph{information distance} (exact equality is generally infeasible over
discrete trajectory indices)---a generalized trajectory distance that
jointly captures kinematic and dynamic variation between points. To formalize this equal-spacing criterion,
we introduce a target distance $D_{\mathrm{target}}$
and cast resampling as an optimization problem:
minimize the total deviation of each segment's information distance
from $D_{\mathrm{target}}$.
The value of $D_{\mathrm{target}}$ controls
the information resolution of the resampled trajectory:
a smaller value retains more points,
while a larger value yields more aggressive compression. The optimization
always enforces comparable information between neighboring retained points,
producing a compact trajectory with $M < N$. In all experiments,
we empirically set a consistent $D_{\mathrm{target}} = 0.05$
as a heuristic compression--fidelity trade-off
across tasks (see Table~\ref{tab:isr_params} in the Appendix).

The key question is how to define this information distance.
Euclidean inter-point distance $\|\vect{p}_{k_m} - \vect{p}_{k_{m+1}}\|$
measures only spatial displacement and is agnostic to acceleration: two segments with vastly different accelerations can span the same spatial gap, yet they carry very different dynamic information. For example, arc-length parameterization
\cite{braglia2025arc} improves spatial uniformity
but remains blind to acceleration-level features:
it treats a straight segment and a sharp turn with the same secant length as equally informative.
To capture both displacement and velocity change
in a single information distance,
we construct an information-intensity field from the velocity
and acceleration norms and use it to warp
the trajectory into an information manifold,
as detailed in the following subsection.

\subsection{Information-Intensity Manifold}

The preceding subsection motivates the need for a unified information distance
that jointly captures kinematic displacement and dynamic velocity variation.
We formalize it through an information-intensity field.

\begin{algorithm}[t]
\caption{Information-Standardized Resampling}
\label{alg:dp_waypoint}
\begin{algorithmic}[1]
\Require Trajectory points $\vect{P} = \{\vect{p}_0, \ldots, \vect{p}_{T-1}\}$,
         $\vect{p}_i \in \mathbb{R}^3$
\Require Timestamps $\vect{t} = \{t_0, \ldots, t_{T-1}\}$
\Require Target distance $D_{\mathrm{target}}$,
         weights $\lambda_{\mathrm{vel}},\, \lambda_{\mathrm{acc}}$
\Ensure  Resampled index set
         $\mathcal{K} \subset \{0, 1, \ldots, T{-}1\}$

\Statex $\triangleright$ \textit{Step 1: Compute velocity and acceleration}
\State $\vect{v}_i \gets (\vect{p}_{i+1} - \vect{p}_i)/\Delta t_i$
       \textbf{for all} $i \in [0,\, T{-}2]$
\State $\vect{a}_i \gets (\vect{v}_i - \vect{v}_{i-1})/\Delta t_i^{\mathrm{avg}}$
       \textbf{for all} $i \in [1,\, T{-}2]$
\State $\vect{a}_0 \gets \vect{a}_1$

\Statex $\triangleright$ \textit{Step 2: Acceleration prefix sum}
\State $S[0] \gets 0$;\;
       $S[i] \gets S[i{-}1] + \|\vect{a}_{i-1}\|$
       \textbf{for} $i = 1, \ldots, T{-}1$

\Statex $\triangleright$ \textit{Step 3: Dynamic programming}
\State $C[0] \gets 0$;\; $\pi[0] \gets -1$
    \Comment{$C$: min cost;\; $\pi$: predecessor}
\For{$i = 1$ \textbf{to} $T{-}1$}
  \State $C[i] \gets +\infty$
  \For{$k = 0$ \textbf{to} $i{-}1$}
    \If{$C[k] + \Call{SegCost}{k,\, i} < C[i]$}
      \State $C[i] \gets C[k] + \Call{SegCost}{k,\, i}$
      \State $\pi[i] \gets k$
    \EndIf
  \EndFor
\EndFor

\Statex $\triangleright$ \textit{Step 4: Backtrack optimal path}
\State $\mathcal{K} \gets \{\,\}$;\;
       $j \gets T{-}1$
\While{$j \neq -1$}
  \State $\mathcal{K} \gets \{j\} \cup \mathcal{K}$
  \State $j \gets \pi[j]$
\EndWhile
\State \Return $\mathcal{K}$

\Statex $\triangleright$ \textit{Segment cost (information-distance deviation)}
\Function{SegCost}{$k,\, i$}
  \State $d_{\mathrm{vel}} \gets \|\vect{p}_i - \vect{p}_k\|$
         \Comment{Secant distance}
  \State $d_{\mathrm{acc}} \gets S[i] - S[k]$
         \Comment{$\sum_{j=k}^{i-1}\|\vect{a}_j\|$}
  \State \Return $(\lambda_{\mathrm{vel}}\, d_{\mathrm{vel}}
                  + \lambda_{\mathrm{acc}}\, d_{\mathrm{acc}}
                  - D_{\mathrm{target}})^2$
\EndFunction
\end{algorithmic}
\end{algorithm}

In Euclidean space, the line element along a trajectory is
\begin{equation}
  ds_{\mathrm{Euc}} = \| \vect{v}(t) \| \, \dd t ,
\end{equation}
which measures pure spatial displacement. To encode both first-order and second-order trajectory content,
we define an \emph{information-intensity field}
\begin{equation}
  \mathcal{L}(t) = \lambda_{\mathrm{vel}}\,\| \vect{v}(t) \|
                  + \lambda_{\mathrm{acc}}\,\| \vect{a}(t) \| ,
\end{equation}
where $\lambda_{\mathrm{vel}}$ and $\lambda_{\mathrm{acc}}$
are non-negative weights that balance the two terms.

We interpret $\mathcal{L}(t)$ as an information density
that warps the trajectory into a one-dimensional manifold $\mathcal{M}$
\cite{blair2002riemannian}:
high-activity regions are stretched,
while low-activity regions are compressed.
The geodesic information distance between trajectory points
at times $t_i$ and $t_j$ is
\begin{equation}
  \Delta s_{\mathcal{M}}(i, j)
    = \int_{t_i}^{t_j} \mathcal{L}(t)\,\dd t .
\end{equation}
Thus, equal geodesic intervals correspond to equal
kinematic--dynamic information content.

\subsection{Geodesic-Equidistant Optimization}

Given the information-intensity manifold $\mathcal{M}$,
we seek a resampled subsequence
$\mathcal{K} = \{k_0, k_1, \ldots, k_{M-1}\}$
such that each segment has geodesic length close to
a prescribed target $D_{\mathrm{target}}$.
For consecutive resampled points $k_m$ and $k_{m+1}$,
where $m = 0, \ldots, M{-}2$,
we approximate the geodesic distance in discrete form as
\begin{equation}
  \Delta s_{\mathcal{M}}
    \approx \lambda_{\mathrm{vel}}\,
            \| \vect{p}_{k_m} - \vect{p}_{k_{m+1}} \|
          + \lambda_{\mathrm{acc}}\,
            \sum_{t=k_m}^{k_{m+1}-1} \| \vect{a}(t) \| ,
  \label{eq:discrete_approx}
\end{equation}
where $\vect{p}$ denotes the end-effector position.
The two terms take deliberately different discrete forms.
The velocity term uses the \emph{secant distance}
$\| \vect{p}_{k_m} - \vect{p}_{k_{m+1}} \|$
rather than the arc length
$\sum_{k_m}^{k_{m+1}} \| \vect{v}(t) \|\,dt$.
Since $\lambda_{\mathrm{vel}}$ aims to equalize
the spatial gaps between consecutive resampled points,
the optimization must directly operate on the quantity
to be equalized---the Euclidean inter-point distance.
If arc length were used instead,
a segment with oscillations or curvature would report
a large path length despite its endpoints being spatially close,
causing the optimizer to incorrectly treat the pair
as well-separated and leaving spatial non-uniformity uncorrected.
By contrast, the acceleration term retains the integral
(summation) form $\sum \| \vect{a}(t) \|$
because it must capture the kinematic--dynamic complexity
\emph{within} each segment rather than between endpoints.
This decoupled design allows $\lambda_{\mathrm{vel}}$
to regulate the spatial distribution of resampled points
while $\lambda_{\mathrm{acc}}$ independently preserves
the operator's dynamic intent.

The local segment cost penalizes deviation
from the target information distance:
\begin{equation}
  \Cost(k_m, k_{m+1})
    = \left[\,
        \Delta s_{\mathcal{M}}(k_m, k_{m+1})
        - D_{\mathrm{target}}
      \,\right]^2 .
\end{equation}

\begin{figure}[t]
  \centering
  \includegraphics[width=\columnwidth]{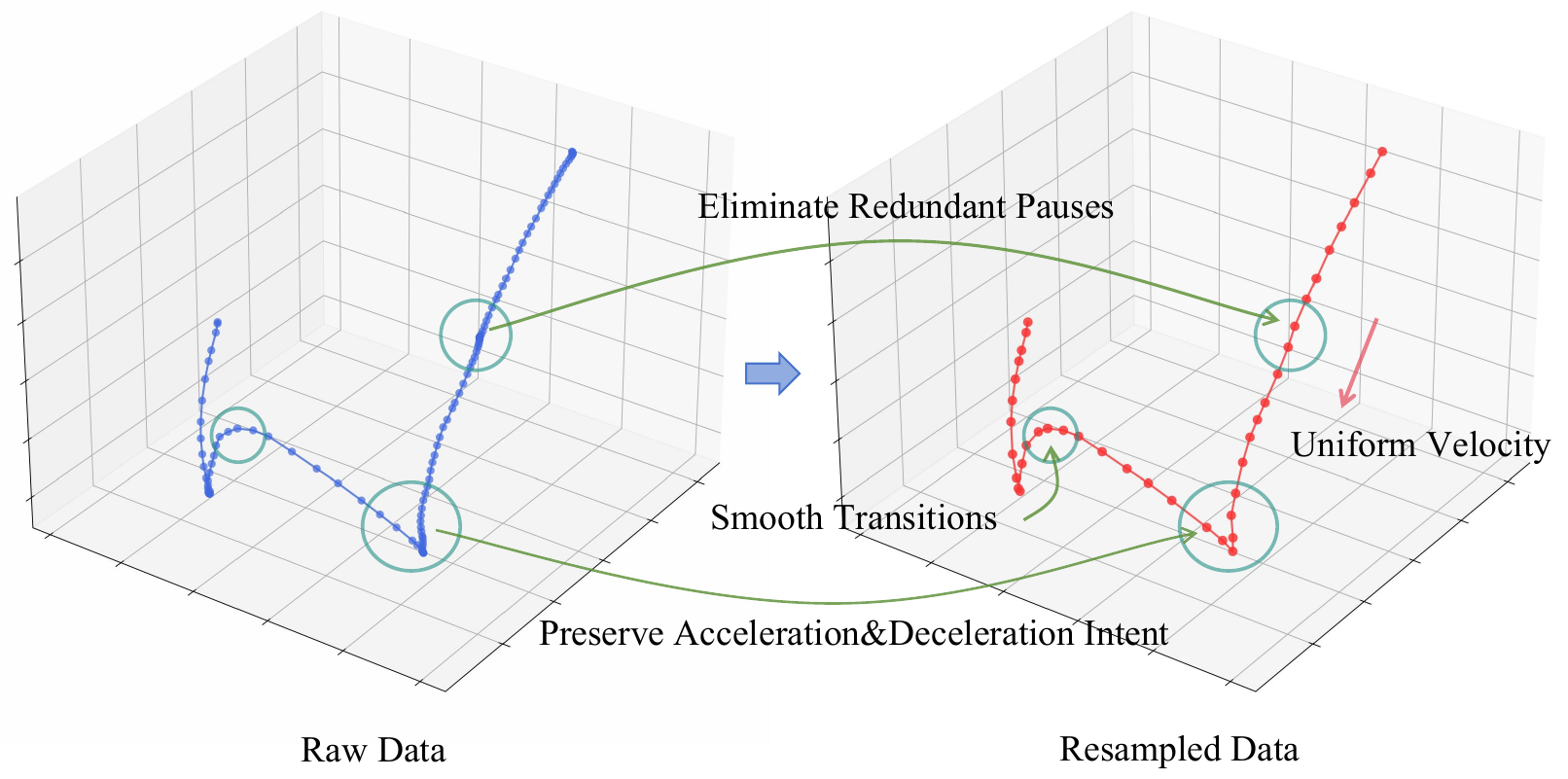}
  \caption{Visualization of ISR resampling effects. Left: raw trajectory segments with redundant pauses and non-uniform velocity. Right: ISR-resampled segments with improved velocity uniformity and preserved high-information phases.}
  \label{fig:visual_traj}
\end{figure}

The global objective over all segments is
\begin{equation}
  J(\mathcal{K})
    = \sum_{m=0}^{M-2} \Cost(k_m, k_{m+1}) .
  \label{eq:global_objective}
\end{equation}
Minimizing $J$ balances compression efficiency with kinematic--dynamic fidelity by assigning larger intervals to low-information segments and finer resolution to high-information segments.
We solve this offline optimization by bottom-up dynamic programming
with $O(T^2)$ time complexity for a length-$T$ trajectory; prefix sums make
each segment-cost evaluation $O(1)$, and pseudocode is given in Algorithm~\ref{alg:dp_waypoint}.
On the acyclic index graph, DP exactly minimizes Eq.~\eqref{eq:global_objective} for fixed weights.
The output trajectory length is adaptive to the accumulated information, which helps retain critical features expressed in end-effector kinematics during compression.
Long trajectories can use banded DP or prune predecessors
whose geodesic distance is far from $D_{\mathrm{target}}$.

\subsection{Effects of the Cost Terms}

We now analyze how the two components
of the discrete geodesic approximation
\eqref{eq:discrete_approx}
shape the resampling behavior
through the global objective $J(\mathcal{K})$
\eqref{eq:global_objective}.

\textbf{Velocity term}.
This term enforces spatial uniformity
between consecutive resampled points,
effectively serving as a denoising mechanism.
Random pauses, hesitations, and slow movements
produce near-zero inter-point distances;
the squared penalty in $J$ drives the optimizer
to merge these into larger, spatially uniform steps.
The result is that the end-effector velocity
across the resampled episode becomes more uniform,
eliminating redundant low-displacement segments
without requiring explicit pause detection.

\textbf{Acceleration term}.
This term preserves trajectory curvature and dynamic intent,
acting in two complementary ways.
First, during arc-like or obstacle-avoidance movements,
directional changes induce centripetal acceleration,
which increases $\mathcal{L}(t)$ locally
and forces denser sampling in high-curvature regions, preventing arcs from being oversimplified
into polygonal approximations.
Second, operators instinctively decelerate
before entering fine manipulation phases
(e.g., aligning and stacking cubes).
Although the spatial displacement during this deceleration window
is small, the tangential acceleration is significant.
The acceleration term amplifies the geodesic distance
in this critical window,
preserving the ``deceleration--preparation--manipulation''
temporal sequence that would otherwise be discarded
as redundant static frames under a purely spatial criterion.

In summary, the velocity term standardizes
the spatial distribution of resampled points
(compression of redundancy),
while the acceleration term standardizes
the preservation of the operator's kinematic--dynamic intent
(retention of critical phases).
Together, the two terms produce the resampling effect
shown in Fig.~\ref{fig:visual_traj}:
the end-effector velocity within each episode becomes more uniform,
redundant pauses are eliminated,
and the operator's manipulation intent is preserved.
This makes the kinematic--dynamic characteristics
across datasets more consistent,
yielding a standardized demonstration distribution
that benefits downstream policy learning.

\begin{figure}[t]
  \centering
  \includegraphics[width=0.80\columnwidth]{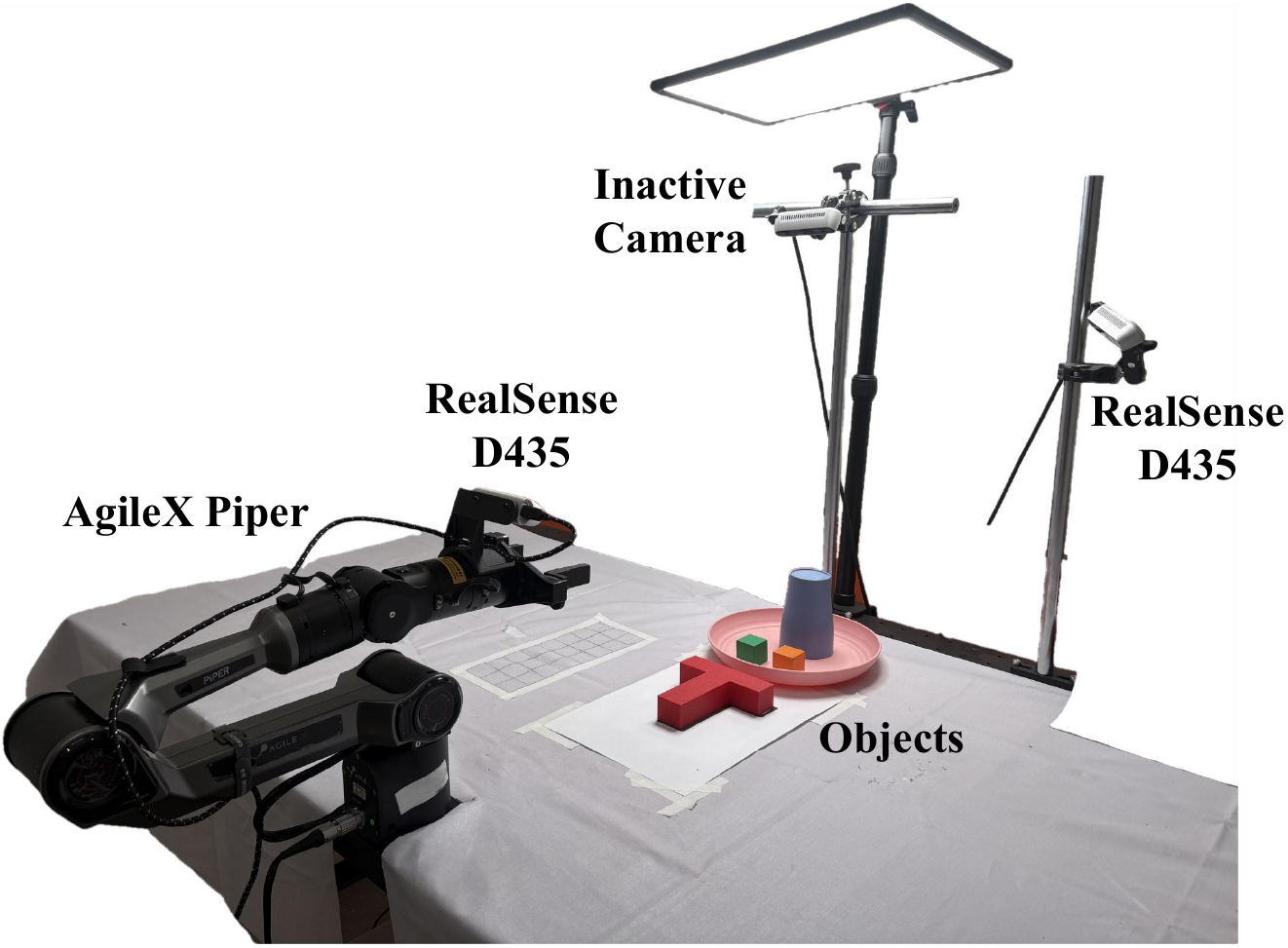}
  \caption{Real-world setup: AgileX Piper arm, two RealSense D435 cameras, and the task objects.}
  \label{fig:task_scene}
\end{figure}

\section{Experiments and Evaluation}

\subsection{Experimental Setup}

We evaluate the effectiveness of ISR-resampled datasets on imitation learning policies across three real-world manipulation tasks.
Fig.~\ref{fig:task_scene} shows our real-world experiment scene and the objects used in the experiments.
We use an AgileX Piper robot arm equipped with a two-finger parallel gripper.
Visual observations are captured by two Intel RealSense D435 cameras: one providing a third-person view RGB stream and the other providing a wrist-mounted RGB stream.
Demonstration datasets are collected via teleoperation using a Synria-Robotics Alicia-D as the leader arm, which controls the Piper follower arm through joint-space mapping \cite{zhao2023learning}.
Each demonstration episode is recorded at 30\,Hz and contains dual camera RGB streams ($(C, H, W) = (3, 480, 640)$ each) along with joint states of both the leader and follower arms.
For IL training, ISR-resampled indices are mapped back to the original episode
to extract the corresponding dual-view RGB frames,
forming a resampled demonstration sequence.

The three real-world tasks are defined as follows:

\textbf{Place\&Cover}: Place the orange cube ($3 \times 3 \times 3$\,cm) into a pink plate (top diameter 22\,cm, height 3\,cm), then pick up the inverted blue cup (bottom diameter 5\,cm, top diameter 7\,cm, height 10.5\,cm) and cover the orange cube.

\textbf{Place\&Stack}: Place the green cube ($3 \times 3 \times 3$\,cm) into the pink plate, then pick up the orange cube and stack it on top of the green cube.

\textbf{Push-T}: Push a red T-shaped cube ($12 \times 12 \times 3$\,cm) into a target region marked by a black outline.

\begin{figure*}[t]
  \centering
  \includegraphics[width=0.87\textwidth]{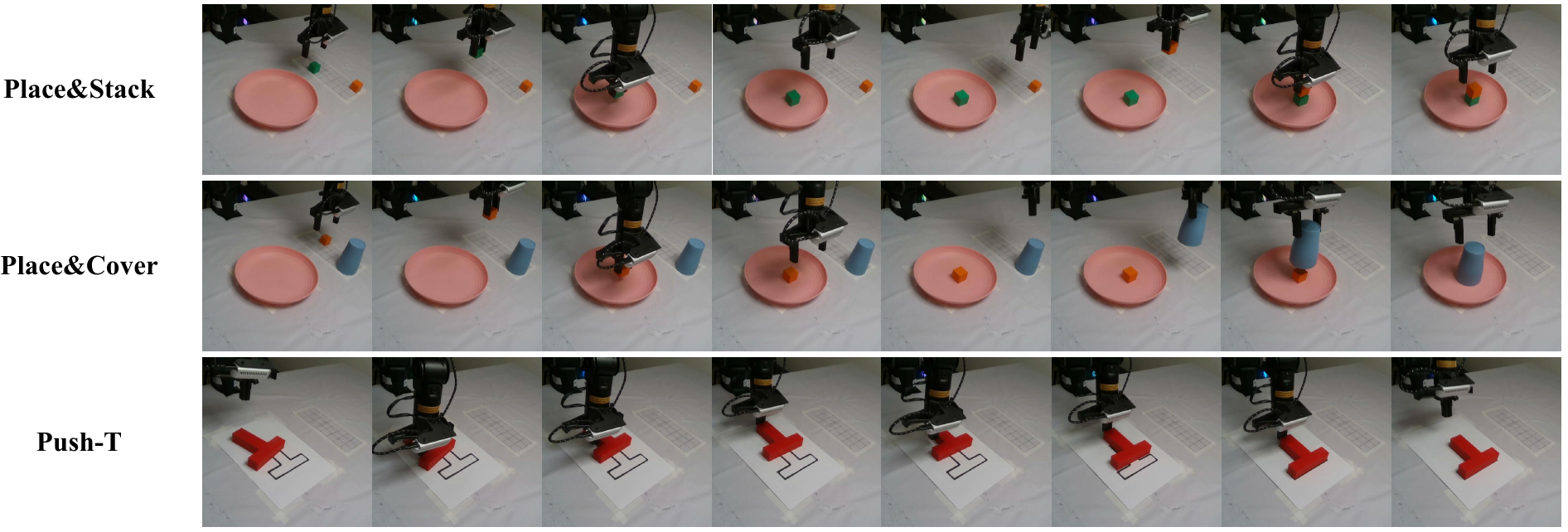}
  \caption{Visualization of the three real-world manipulation tasks: Place\&Cover, Place\&Stack, and Push-T.}
  \label{fig:experiment_tasks}
\end{figure*}

The manipulation procedures are illustrated in Fig.~\ref{fig:experiment_tasks}.

\begin{figure}[t]
  \centering
  \includegraphics[width=0.80\columnwidth]{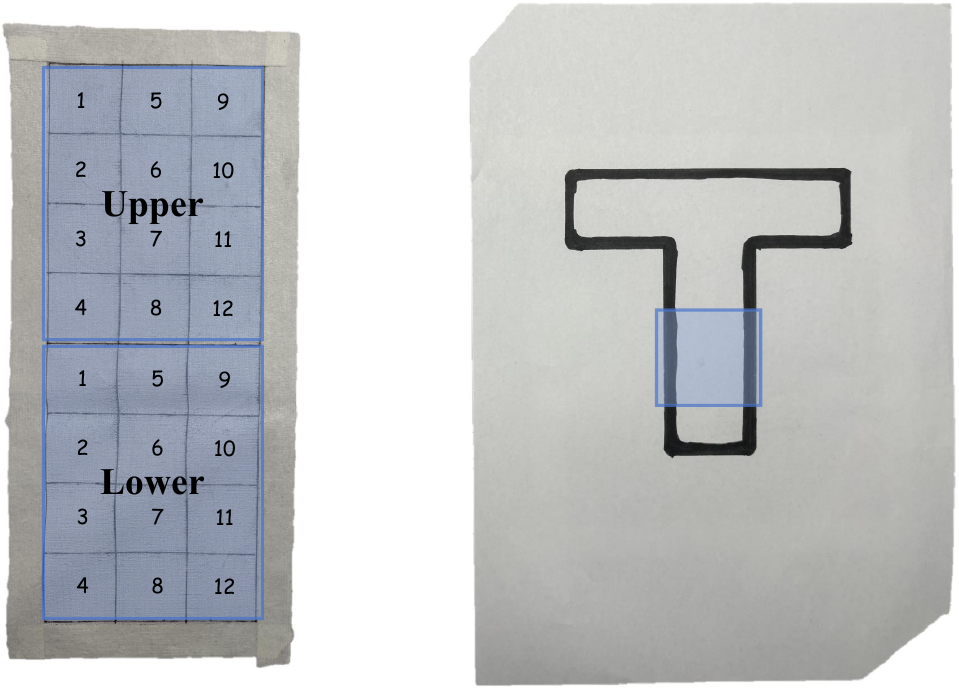}
  \caption{Manipulation areas. Left: $3 \times 8$ grid cells (each $3 \times 3$\,cm) for object placement in Place\&Stack and Place\&Cover. Right: target region (black outline) for Push-T.}
  \label{fig:manipulation_area}
\end{figure}

We collect 150 demonstration episodes for Place\&Stack and Push-T, and 100 episodes for Place\&Cover.
For the Place\&Stack and Place\&Cover tasks, the workspace is uniformly divided into a 24 cells, as shown in Fig.~\ref{fig:manipulation_area}.
In each episode, the two objects are randomly placed in distinct cells via combinatorial sampling.
Specifically, in Place\&Stack, the green cube is randomly placed in one of the lower 12 cells and the orange cube in one of the upper 12 cells.
In Place\&Cover, the orange cube is randomly placed in one of the lower 12 cells and the blue cup in one of the upper 12 cells, where each cube occupies one cell and each cup occupies four cells.
For the Push-T task, a red T-shaped cube is placed with random positional offsets and rotations relative to its target pose, as shown in Fig.~\ref{fig:manipulation_area}.
The operator then pushes the T-shaped cube into the target region without adhering to any prescribed manipulation sequence, allowing natural and varied pushing strategies.

\subsection{Experiment Design}
\textbf{Baseline Comparison}: We select time-uniform $3\times$ downsampling as our evaluation baseline and compare it against ISR across all three real-world tasks using two imitation learning policies: $\pi_{0.5}$ \cite{intelligence2025pi_} and VO-DP \cite{ni2025vodp}. Training hyperparameters and codebase references are provided in Table~\ref{tab:policy_params}(Appendix); all other settings follow the default configurations of the respective codebases.

\textbf{Ablation Study}: We ablate the acceleration weight $\lambda_{\mathrm{acc}}$ while keeping $\lambda_{\mathrm{vel}}$ fixed. The velocity term provides the fundamental spatial denoising that every resampled trajectory requires; removing it would collapse the method to a purely acceleration-based criterion that no longer enforces spatial uniformity. The acceleration term is equally indispensable: in preliminary trials with $\lambda_{\mathrm{acc}} = 0$, the optimizer, driven solely by spatial cost, skips temporally dense but spatially compact phases---for example, reducing a ``descend--grasp--lift'' sequence to a direct descend-to-lift shortcut and discarding the grasp action entirely, which leads to severe policy degradation. The practical question is how to set the acceleration weight. Because the optimal $\lambda_{\mathrm{acc}}$ depends on the dynamic complexity of each task, we vary it across values to study its effect on downstream performance.

\textbf{Cross-Operator Robustness}: To evaluate robustness to operator-specific demonstration styles, we construct a mixed-operator dataset for Place\&Stack by combining demonstrations from three different operators, each contributing 50 episodes (150 episodes in total). All operators follow a unified object randomization protocol as described above. We then train $\pi_{0.5}$ on this mixed dataset and compare ISR against the time-uniform downsampling baseline.

\begin{table}[t]
  \centering
  \caption{Baseline comparison. Success rate (\%) with resampling ratio (\%) in parentheses.}
  \label{tab:baseline_comparison}
  \setlength{\tabcolsep}{5pt}
  \renewcommand{\arraystretch}{1.3}
  \begin{tabular}{lcccc}
    \hline
    \textbf{Method}
      & \textbf{Place\&Cover}
      & \textbf{Place\&Stack}
      & \textbf{Push-T}
      & \textbf{AVG.($\uparrow$)} \\
    \hline
    3$\times$ + $\pi_{0.5}$
      & 45.8\,{\scriptsize(33.3)}
      & 48.6\,{\scriptsize(33.3)}
      & 48.9\,{\scriptsize(33.3)}
      & 47.8 \\
    \rowcolor{blue!12}
    ISR + $\pi_{0.5}$
      & 72.2\,{\scriptsize(28.9)}
      & 72.2\,{\scriptsize(28.2)}
      & 71.1\,{\scriptsize(40.2)}
      & \textbf{71.8} \\
    3$\times$ + VO-DP
      & 56.9\,{\scriptsize(33.3)}
      & 66.7\,{\scriptsize(33.3)}
      & 64.4\,{\scriptsize(33.3)}
      & 62.7 \\
    \rowcolor{blue!12}
    ISR + VO-DP
      & 77.8\,{\scriptsize(28.9)}
      & 91.6\,{\scriptsize(28.2)}
      & 71.1\,{\scriptsize(40.2)}
      & \textbf{80.2} \\
    \hline
  \end{tabular}
\end{table}

\subsection{Evaluation Protocol}

Each task is evaluated with a fixed set of object configurations
that systematically cover the workspace (Fig.~\ref{fig:manipulation_area}).

\textbf{Place\&Cover} (72 trials).
The blue cup traverses 6 positions in the upper 12 cells,
each occupying four adjacent cells:
$\{1,2,5,6\}$, $\{2,3,6,7\}$, $\{3,4,7,8\}$,
$\{5,6,9,10\}$, $\{6,7,10,11\}$, $\{7,8,11,12\}$.
For each cup position, the orange cube is placed
in each of the 12 lower cells once,
yielding $6 \times 12 = 72$ trials in total.

\textbf{Place\&Stack} (72 trials).
The orange cube traverses 6 fixed positions
in the upper 12 cells: $\{1, 3, 6, 8, 9, 11\}$.
For each orange-cube position,
the green cube is placed in each of the 12 lower cells once,
yielding $6 \times 12 = 72$ trials.

\textbf{Push-T} (45 trials).
The center of the T-shaped cube is placed
at 15 uniformly distributed positions within the blue region
of Fig.~\ref{fig:manipulation_area}.
At each position, 3 initial orientations are tested:
$-20^{\circ}$, $0^{\circ}$, and $+20^{\circ}$,
yielding $15 \times 3 = 45$ trials.

For all tasks, the success rate is computed as the ratio
of successful trials to the total number of trials.

\subsection{Real-world Performance}

The ISR settings in Table~\ref{tab:isr_params} (Appendix) also serve as a cross-task heuristic: fix spatial settings and scale acceleration weight with contact dynamics.

\begin{figure}[t]
  \centering
  \includegraphics[width=0.85\columnwidth]{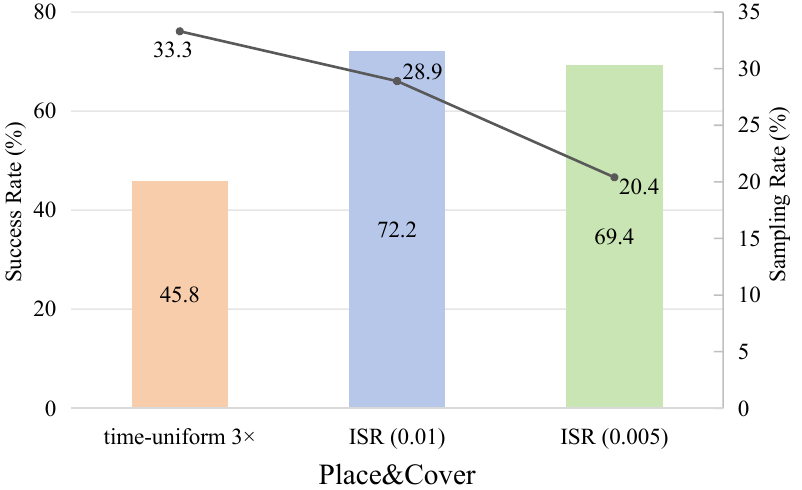}\\[4pt]
  \includegraphics[width=0.85\columnwidth]{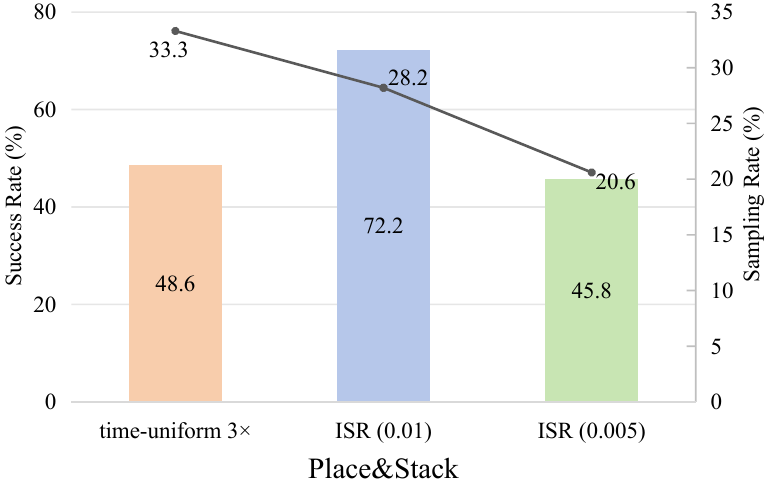}
  \caption{Effect of the acceleration weight $\lambda_{\mathrm{acc}}$ on task success rate and resampling ratio. Top: Place\&Cover. Bottom: Place\&Stack.}
  \label{fig:ablation}
\end{figure}

\begin{figure}[h]
  \centering
  \includegraphics[width=\columnwidth]{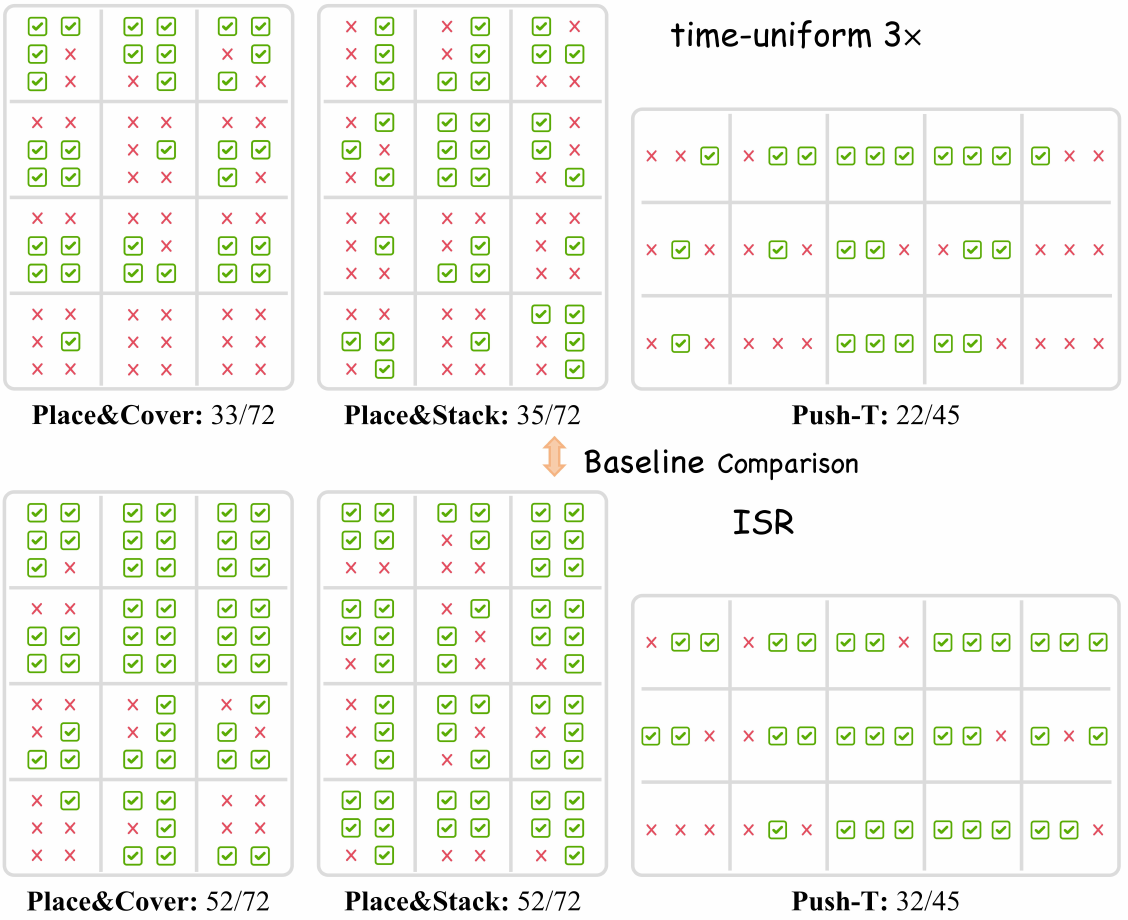}
  \caption{Per-trial evaluation results of $\pi_{0.5}$ on the three real-world tasks.
  A checkmark (\checkmark) denotes a successful trial and a cross ($\times$) denotes a failure.
  Following the evaluation protocol, each large cell corresponds to one of the 12 positions
  traversed by the orange cube (Place\&Cover) and the green cube (Place\&Stack).
  Within each large cell, the 6 small entries represent the paired traversal positions
  of the blue cup (Place\&Cover) or the orange cube (Place\&Stack).}
  \label{fig:result_detail}
\end{figure}

\textbf{Baseline Comparison}.
Table~\ref{tab:baseline_comparison} summarizes the task success rates
of ISR versus time-uniform $3\times$ downsampling
across three tasks and two policies.
With $\pi_{0.5}$, ISR raises the average success rate
from 47.8\% to 71.8\%, an absolute improvement of 24.0 percentage points; all three tasks benefit consistently.
The same trend holds for VO-DP, where ISR consistently outperforms
the time-uniform baseline across all evaluated tasks.
Notably, ISR achieves these gains while using comparable
or fewer training samples---the resampling ratios
in Table~\ref{tab:baseline_comparison} show that ISR retains
28--40\% of the original action points,
similar to or less than the fixed 33.3\% of $3\times$ downsampling.
These results demonstrate that the gains stem from
the improved information distribution of ISR-resampled sequences,
not from retaining more data; per-trial $\pi_{0.5}$ results
are provided in Fig.~\ref{fig:result_detail} (Appendix).

\begin{figure}[t]
  \centering
  \includegraphics[width=0.81\columnwidth]{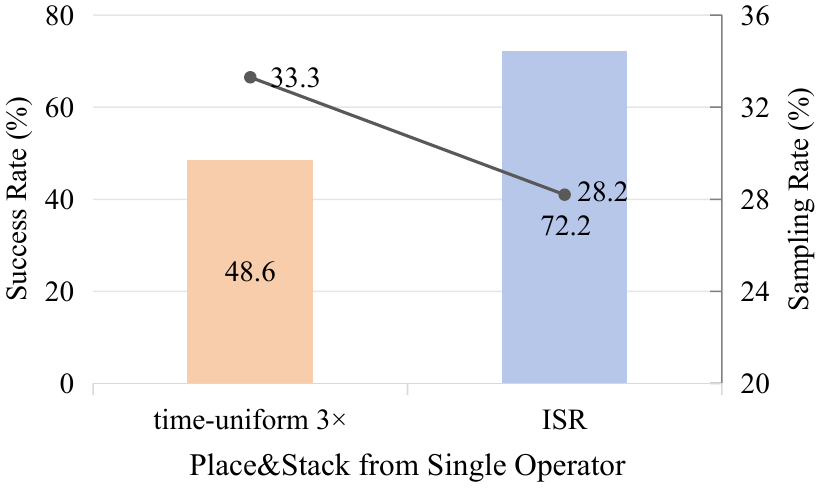}\\[4pt]
  \includegraphics[width=0.81\columnwidth]{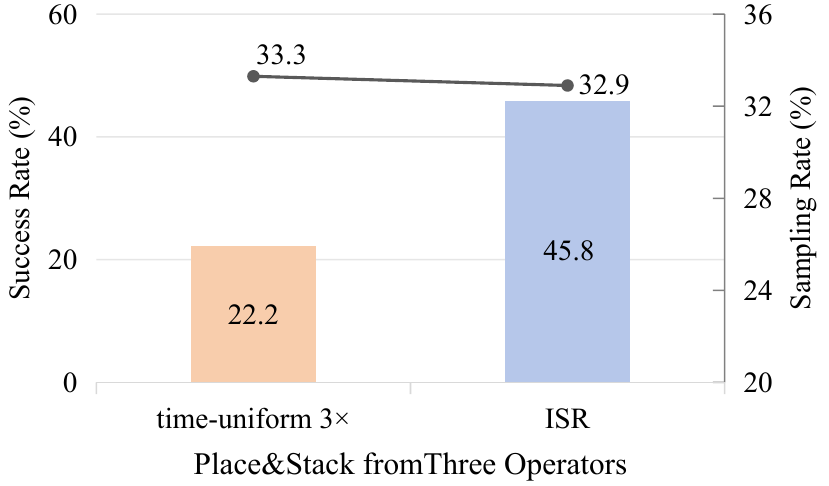}
  \caption{Cross-operator robustness on Place\&Stack. Top: single-operator dataset (150 episodes). Bottom: mixed-operator dataset (3 operators, 50 episodes each). }
  \label{fig:across_operator}
\end{figure}

\textbf{Ablation Study}.
We train $\pi_{0.5}$ on Place\&Cover and Place\&Stack
with ISR under two $\lambda_{\mathrm{acc}}$ settings (0.01 and 0.005),
keeping $\lambda_{\mathrm{vel}} = 1.0$
and $D_{\mathrm{target}} = 0.05$ fixed.
Results are shown in Fig.~\ref{fig:ablation}.
Reducing $\lambda_{\mathrm{acc}}$ produces two coupled effects:
(i)~the acceleration term contributes less to the geodesic distance,
so segments with high dynamic variation are no longer stretched
on the information manifold;
(ii)~the resulting trajectories are compressed more aggressively,
as the sampling rate drops from approximately 28\% to 20\%.
The impact on task performance, however, is task-dependent.
On Place\&Cover, the success rate decreases only marginally
(72.2\% $\to$ 69.4\%),
because covering an object is geometrically forgiving
and does not rely heavily on precise deceleration control.
On Place\&Stack, success drops sharply
(72.2\% $\to$ 45.8\%), returning to the time-uniform baseline.
Unlike covering, stacking requires controlled deceleration-to-contact;
when $\lambda_{\mathrm{acc}}$ is too small, the critical
``decelerate--align--place'' phase is under-sampled,
so the policy loses force-sensitive intent.

These results confirm that the acceleration term
is indispensable for tasks involving contact-rich manipulation,
not an optional addition that merely fine-tunes performance.
Moderate $\lambda_{\mathrm{acc}}$ values therefore form a stable operating range,
preserving high-acceleration phases without substantially increasing
the resampling ratio.
Conversely, an excessively large $\lambda_{\mathrm{acc}}$
would over-stretch acceleration-rich segments, retain too many action points,
and weaken the spatial uniformity enforced by the velocity term.
Thus, $\lambda_{\mathrm{acc}}$ should match each task's dynamic complexity,
balancing compression efficiency against dynamic fidelity.

\textbf{Cross-Operator Robustness}.
To evaluate robustness to operator-specific demonstration styles,
we train $\pi_{0.5}$ on Place\&Stack using two dataset configurations:
one collected by a single operator
and one combining episodes from different operators,
both following the same object randomization protocol.
Results are shown in Fig.~\ref{fig:across_operator}.
With a single operator,
ISR improves success from 48.6\% to 72.2\%
(Table~\ref{tab:baseline_comparison}).
When demonstrations from different operators are mixed,
time-uniform $3\times$ downsampling suffers a severe performance collapse:
success drops from 48.6\% to 22.2\%,
because operator-specific speed, pause patterns,
and stylistic differences are preserved in the downsampled data,
introducing conflicting training signals.
Under the same mixed-operator setting,
ISR achieves 45.8\%---more than doubling
the $3\times$ result (22.2\%)
and approaching the single-operator $3\times$ level (48.6\%).
These results demonstrate that ISR effectively absorbs
inter-operator variability through kinematic--dynamic standardization,
making it a practical solution for scaling demonstration collection
across multiple operators.


\section{Conclusions}

We presented ISR, a data-centric trajectory resampling method
that standardizes teleoperated demonstrations
by enforcing approximately equal kinematic--dynamic information distance
between consecutive action points.
A velocity term compresses redundant pauses and hesitations,
while an acceleration term preserves sampling density
through high-curvature and deceleration-to-contact phases.
Together they define an information distance
under which geodesic-equidistant optimization
produces compact, information-uniform trajectories.
Across three real-world tasks,
ISR raises the average success rate
from 47.8\% to 71.8\% with $\pi_{0.5}$
and from 62.7\% to 80.2\% with VO-DP,
while retaining comparable or fewer action points
than time-uniform $3\times$ downsampling.
Ablation confirms that the acceleration term is essential
for force-sensitive tasks such as cube stacking,
and cross-operator experiments show that ISR
effectively absorbs inter-operator variability.

\textbf{Limitations.}
ISR currently operates only on end-effector position trajectories;
extending the information-intensity field
to joint-space representations
could broaden its applicability.
Additionally, the acceleration weight $\lambda_{\mathrm{acc}}$
requires task-specific manual tuning,
as different manipulation tasks exhibit varying sensitivity
to acceleration-level features.
An adaptive mechanism that infers this weight
from trajectory statistics would improve generality.



\section*{Appendix}
\begin{table}[h]
  \centering
  \caption{Policy training parameters.
           All policies share the same configuration.}
  \label{tab:policy_params}
  \setlength{\tabcolsep}{8pt}
  \renewcommand{\arraystretch}{1.3}
  \begin{tabular}{lcc}
    \hline
    \textbf{Parameter} & \textbf{$\pi_{0.5}$} & \textbf{VO-DP}\\
    \hline
    Batch size              & 128 & 128\\
    Action chunk horizon    & 8 & 8\\
    Observation history length & -- & 1\\
    Training iterations     & 20000 & 20000\\
    Third-person camera shape     &  CHW = (3, 224, 224)  &  (3, 480, 640) \\
    Wrist camera shape            & CHW = (3, 224, 224) & -- \\
    \hline
  \end{tabular}
\end{table}

We use two open-source policy implementations.
$\pi_{0.5}$ is based on the openpi codebase\footnote{\url{https://github.com/Physical-Intelligence/openpi}},
and VO-DP is based on the DRRM codebase\footnote{\url{https://github.com/D-Robotics-AI-Lab/DRRM}}.
Table~\ref{tab:policy_params} lists the training hyperparameters; all remaining configurations follow the respective defaults.

\begin{table}[h]
  \centering
  \caption{ISR resampling parameters used in all experiments
           (except the ablation study).}
  \label{tab:isr_params}
  \setlength{\tabcolsep}{8pt}
  \renewcommand{\arraystretch}{1.3}
  \begin{tabular}{lccc}
    \hline
    \textbf{Task}
      & $D_{\mathrm{target}}$
      & $\lambda_{\mathrm{vel}}$
      & $\lambda_{\mathrm{acc}}$ \\
    \hline
    Place\&Cover  & 0.05 & 1.0 & 0.01 \\
    Place\&Stack  & 0.05 & 1.0 & 0.01 \\
    Push-T        & 0.05 & 1.0 & 0.03 \\
    \hline
  \end{tabular}
\end{table}

\bibliographystyle{IEEEtran}
\bibliography{references}

\end{document}